\documentclass[10pt, a4paper]{article} 

\usepackage{arxiv}
\usepackage[utf8]{inputenc}
\usepackage[T1]{fontenc}
\usepackage{hyperref}
\usepackage{url}
\usepackage{booktabs}
\usepackage{amsfonts}
\usepackage{amsmath}
\usepackage{nicefrac}
\usepackage{microtype}
\usepackage{cleveref}
\usepackage{graphicx}
\usepackage{doi}
\usepackage{tabularx}
\usepackage{tikz}
\usetikzlibrary{shapes.geometric, arrows, positioning}
\usepackage[english]{babel}
\usepackage{multicol} 
\usepackage{caption}
\usepackage[compact]{titlesec} 
\usepackage{enumitem} 
\setlist{nosep}

\renewcommand{\headeright}{}
\renewcommand{\undertitle}{}

\title{Sentiment Analysis and Customer Satisfaction Prediction on E-Commerce Platforms Based on YouTube Comments Using the XGBoost Algorithm}

\author{%
  Ridho Benedictus Togi Manik\\
  Department of Data Science\\
  Institut Teknologi Sumatera \\
  South Lampung, 35365, Indonesia \\
  \texttt{ridho.123450060@student.itera.ac.id}
  \And
  Muhammad Aqil Ramadhan\\
  Department of Data Science\\
  Institut Teknologi Sumatera\\
  Lampung Selatan, 35365, Indonesia\\
  \texttt{muhammad.123450066@student.itera.ac.id}
  \And
  Ihsan Maulana Yusuf\\
  Department of Data Science\\
  Institut Teknologi Sumatera\\
  Lampung Selatan, 35365, Indonesia\\
  \texttt{ihsan.123450110@student.itera.ac.id}
  \And
  Luluk Muthoharoh, S.Si., M.Si. \\
  Department of Data Science\\
  Institut Teknologi Sumatera \\
  Lampung Selatan, Indonesia \\
  \texttt{luluk.muthoharoh@sd.itera.ac.id}
  \And
  Ardika Satria, S.Si., M.Si. \\
  Department of Data Science\\
  Institut Teknologi Sumatera \\
  Lampung Selatan, Indonesia \\
  \texttt{ardika.satria@sd.itera.ac.id}
  \And
  Martin Clinton Tosima Manullang, Ph.D.\\
  Department of Informatics Engineering\\
  Institut Teknologi Sumatera\\
  Lampung Selatan, 35365, Indonesia\\
  \texttt{martin.manullang@if.itera.ac.id}\\
}

\renewcommand{\shorttitle}{Sentiment Analysis of E-Commerce via YouTube Comments}

\begin{document}
\maketitle

% ── Abstract ───────────────────────────────────────────────────────────────────
\vspace{-0.5cm}
\begin{abstract}
\textit{The exponential expansion of digital commerce in Indonesia has significantly shifted consumer interactions toward video-centric social networks, particularly YouTube. Consequently, the sheer volume of unstructured, multi-contextual comments poses a tremendous challenge for manual sentiment tracking. This study investigates and constructs a predictive model for customer satisfaction leveraging the Extreme Gradient Boosting (XGBoost) architecture coupled with Term Frequency-Inverse Document Frequency (TF-IDF) vectorization. By utilizing a secondary dataset of YouTube comments retrieved from e-commerce review videos, the raw text underwent rigorous preprocessing to generate normalized numerical features. The experimental results demonstrate that the PyCaret-optimized Machine Learning framework delivers superior classification resilience. Beyond standard performance metrics, profound lexical evaluations and feature importance mapping uncover a unique phenomenon: the e-commerce discourse is heavily infiltrated by socio-political terminologies, which ultimately dictate the polarity of audience satisfaction.}
\end{abstract}


\vspace{0.3cm}

% ── MULTI-COLUMN START ─────────────────────────────────────────────────────────
\begin{multicols}{2}

% ── 1. INTRODUCTION ────────────────────────────────────────────────────────────
\section{Introduction}
Integrating predictive modeling within the e-commerce sector has evolved from an optional advantage to an absolute necessity. Modern consumers no longer restrict their grievances and feedback strictly to official application storefronts; instead, they vocalize their experiences dynamically on public platforms such as YouTube. The resulting textual data is not only massive in scale but heavily unstructured, making conventional manual observation virtually impossible \cite{jurnal1}. Users frequently compose hybrid reviews where positive affirmations and critical complaints coexist within the same paragraph, necessitating sophisticated data mining techniques to properly isolate and measure satisfaction levels \cite{jurnal2}.

A multitude of scholarly works has attempted to address the complexities of public opinion classification. However, the continuous influx of noisy YouTube interactions remains a formidable obstacle for corporate evaluation \cite{jurnal4}. Traditional linear models frequently fall short in deciphering the non-linear semantic relationships embedded within internet slang and informal dialogues that govern social media ecosystems \cite{jurnal5}.

To circumvent the limitations of rudimentary classifiers, this research proposes the implementation of Extreme Gradient Boosting (XGBoost). Operating as a highly advanced iteration of tree-based algorithms, XGBoost is inherently designed to process high-dimensional text data, effectively mitigate overfitting, and yield exceptional precision compared to legacy methodologies \cite{jurnal6}. Consequently, this paper intends to rigorously assess the predictive capabilities of XGBoost on YouTube-derived e-commerce sentiments and directly benchmark its efficacy against Deep Learning counterparts, specifically Long Short-Term Memory (LSTM) networks. The insights generated from this evaluation aim to facilitate a more responsive, real-time public sentiment monitoring framework for digital enterprises.

% ── 2. LITERATURE REVIEW ───────────────────────────────────────────────────────
\section{Literature Review}

\subsection{Previous Research}
Establishing a robust academic foundation is imperative to identifying existing knowledge gaps and selecting optimal computational parameters. A condensed overview of preceding scholarly investigations is detailed in Table \ref{tab:previous_research}.

\vspace{0.3cm}
\begin{center}
    \captionof{table}{Summary of Prior Investigations}
    \label{tab:previous_research}
    \scriptsize 
    \renewcommand{\arraystretch}{1.1} 
    \begin{tabularx}{\columnwidth}{@{} >{\raggedright\arraybackslash}p{1.5cm} >{\centering\arraybackslash}p{0.8cm} >{\raggedright\arraybackslash}p{2.6cm} >{\raggedright\arraybackslash}X @{}}
        \toprule
        \textbf{Author} & \textbf{Year} & \textbf{Focus \& Method} & \textbf{Key Findings} \\
        \midrule
        Daza et al. & 2024 & E-Commerce Reviews (SLR) & SVM \& LSTM proved superior \cite{jurnal1}. \\
        Ramadhani & 2025 & Lazada Satisfaction (C4.5) & Acc: 77.85\% \cite{jurnal2}. \\
        Sondakh & 2024 & Shopee Play Store (SVM) & Acc: 90.8\% \cite{jurnal3}. \\
        Darmawan & 2022 & Tokopedia Polarity (NB) & High processing efficiency \cite{jurnal4}. \\
        \bottomrule
    \end{tabularx}
\end{center}
\vspace{0.3cm}

\subsection{E-Commerce Sentiment Dynamics}
The discipline of sentiment analysis—a subfield of Natural Language Processing (NLP)—focuses on decoding the emotional polarity encapsulated within user-generated text \cite{jurnal7}. As video platforms become the primary source of product referencing, parsing YouTube comments introduces distinct hurdles, primarily due to grammatical inconsistencies, typographical errors, and regional slang \cite{jurnal4}. Combining meticulous feature extraction with robust statistical algorithms is therefore non-negotiable for accurate sentiment prediction.

\subsection{Extreme Gradient Boosting (XGBoost)}
Functioning as an optimized implementation of the Gradient Boosting Decision Tree (GBDT) framework, XGBoost is renowned for its computational agility and scalability \cite{jurnal6}. The algorithm learns iteratively by constructing subsequent decision trees intended solely to minimize the residual errors generated by prior iterations. Its objective function is mathematically denoted as follows \cite{chen2016}:
\begin{equation}
\mathcal{L}^{(t)} = \sum_{i=1}^n l(y_i, \hat{y}_i^{(t-1)} + f_t(x_i)) + \Omega(f_t)
\end{equation}

% ── 3. RESEARCH METHODOLOGY ────────────────────────────────────────────────────
\section{Research Methodology}

\subsection{Data Acquisition and Processing Strategy}
Employing a quantitative, data-centric framework, this study relies on secondary information harvested from YouTube comments targeting Indonesian e-commerce review content \cite{jurnal2, jurnal5}.

The analytical pipeline commences with Text Preprocessing—encompassing case folding, punctuation cleansing, tokenization, stopword elimination, and stemming—to aggressively filter out conversational noise \cite{jurnal9}. Subsequently, Feature Engineering translates the purified textual corpus into measurable vectors via the Term Frequency-Inverse Document Frequency (TF-IDF) paradigm \cite{jurnal8}. To guard against model memorization (overfitting), the corpus is partitioned into training and validation sets using an 80:20 ratio. The procedural sequence is visualized in Figure \ref{fig:diagram}.

\vspace{0.2cm}
\begin{center}
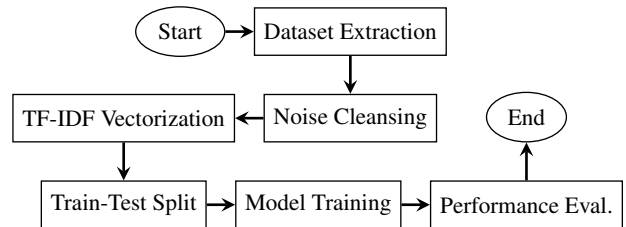

\resizebox{\columnwidth}{!}{%
\begin{tikzpicture}[
    node distance=0.8cm and 0.4cm, 
    base/.style={draw, align=center, minimum height=0.5cm, font=\scriptsize, fill=white, inner sep=0.1cm},
    startstop/.style={base, ellipse},
    process/.style={base, rectangle}, 
    arrow/.style={thick,->,>=stealth}
]
    \node (mulai) [startstop] {Start};
    \node (kumpul) [process, right=0.3cm of mulai] {Dataset Extraction};
    \node (prepro) [process, below=0.4cm of kumpul] {Noise Cleansing};
    \node (tfidf) [process, left=0.3cm of prepro] {TF-IDF Vectorization};
    \node (split) [process, below=0.4cm of tfidf] {Train-Test Split};
    \node (xgb) [process, right=0.3cm of split] {Model Training};
    \node (eval) [process, right=0.3cm of xgb] {Performance Eval.};
    \node (selesai) [startstop, above=0.4cm of eval] {End};

    \draw [arrow] (mulai) -- (kumpul); \draw [arrow] (kumpul) -- (prepro);
    \draw [arrow] (prepro) -- (tfidf); \draw [arrow] (tfidf) -- (split);
    \draw [arrow] (split) -- (xgb); \draw [arrow] (xgb) -- (eval); \draw [arrow] (eval) -- (selesai);
\end{tikzpicture}%
}
\captionof{figure}{Methodological Architecture}
\label{fig:diagram}
\end{center}
\vspace{0.2cm}

% ── 4. RESULTS AND DISCUSSION ──────────────────────────────────────────────────
\section{Results and Discussion}

\subsection{Exploratory Data Analysis (EDA)}

\subsubsection{Sentiment and Emotion Dispersion}
An initial examination of the sentiment labels exposes a drastically skewed distribution (Figure \ref{fig:sentiment_dist}). The conversational landscape is overwhelmingly negative, representing 63.2\% of the dataset. Neutral and positive remarks constitute merely 30.0\% and 6.8\%, respectively. 

\vspace{0.2cm}
\begin{center}
    \includegraphics[width=\columnwidth]{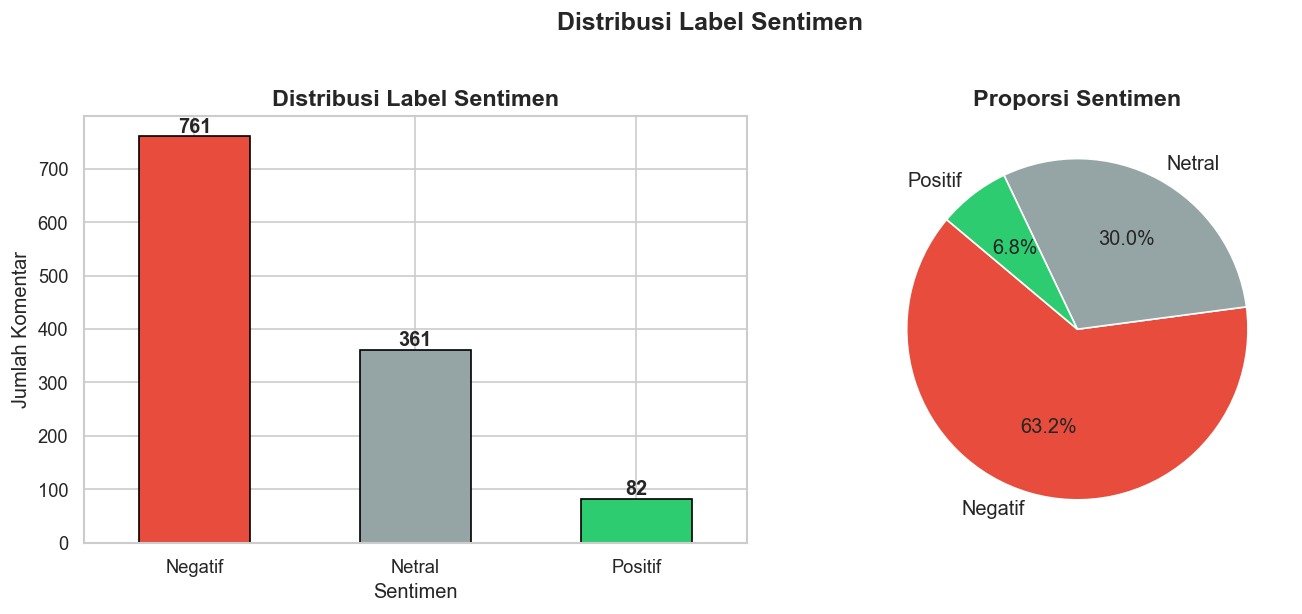} 
    \captionof{figure}{Proportional Distribution of Sentiment Labels}
    \label{fig:sentiment_dist}
\end{center}
\vspace{0.2cm}

To acquire a more granular psychological perspective, the macroscopic sentiments were categorized into definitive emotions. The dominant negativity is fundamentally fueled by intense Disappointment (\textit{Kecewa}) and Anger/Hate (\textit{Marah/Benci}). This severe class disparity strongly dictates the requirement for resilient algorithms capable of mitigating imbalanced learning conditions.

\vspace{0.2cm}
\begin{center}
    \includegraphics[width=\columnwidth]{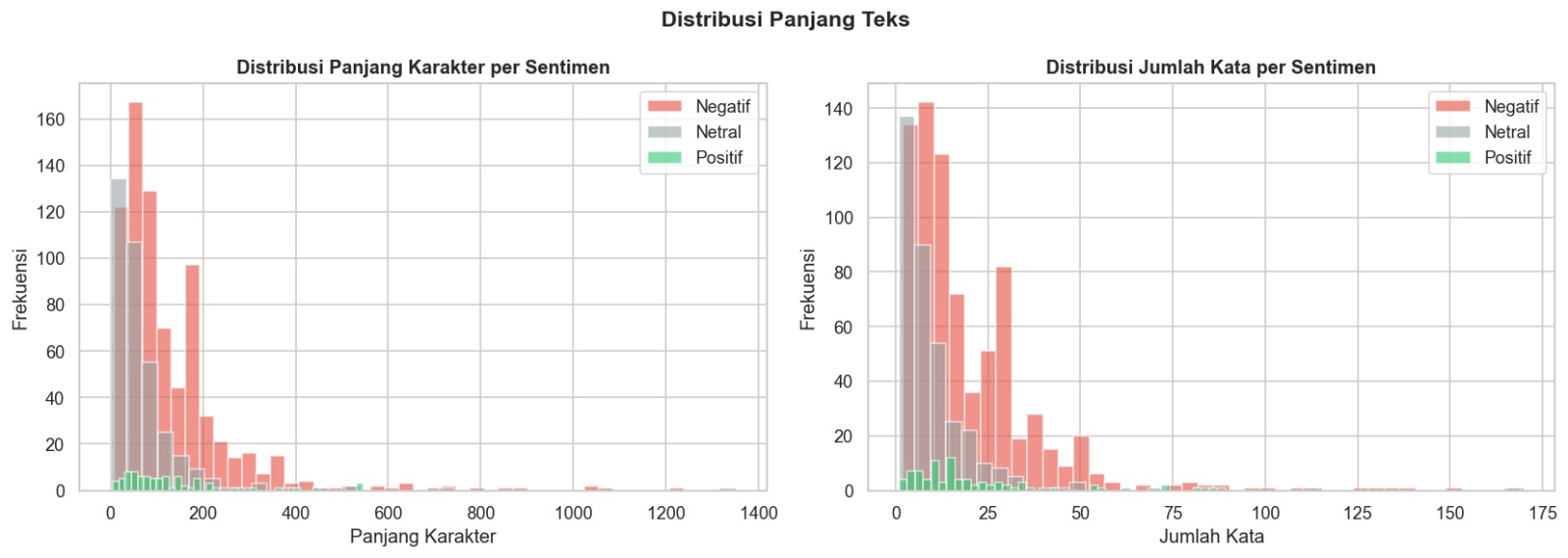}\\
    \vspace{0.1cm}
    \includegraphics[width=\columnwidth]{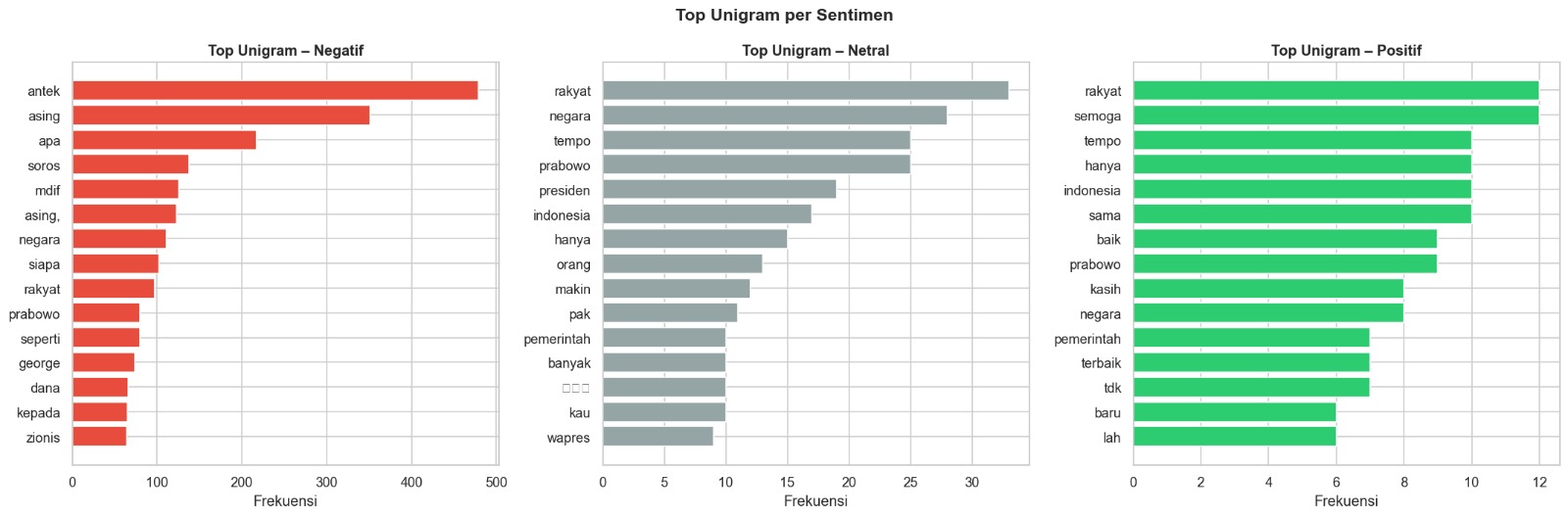}
    \captionof{figure}{Emotion Categories (Top) and Emotion-Sentiment Crosstabulation (Bottom)}
    \label{fig:emotions_combined}
\end{center}
\vspace{0.2cm}

\subsubsection{Textual Volume Characteristics}
Behavioral patterns become evident when analyzing comment lengths. As illustrated in Figure \ref{fig:text_length}, negative remarks not only exhibit broader variance but are noticeably longer than positive statements. This suggests a tendency for frustrated audiences to articulate their grievances through extensive, paragraph-format arguments.

\vspace{0.2cm}
\begin{center}
    \includegraphics[width=\columnwidth]{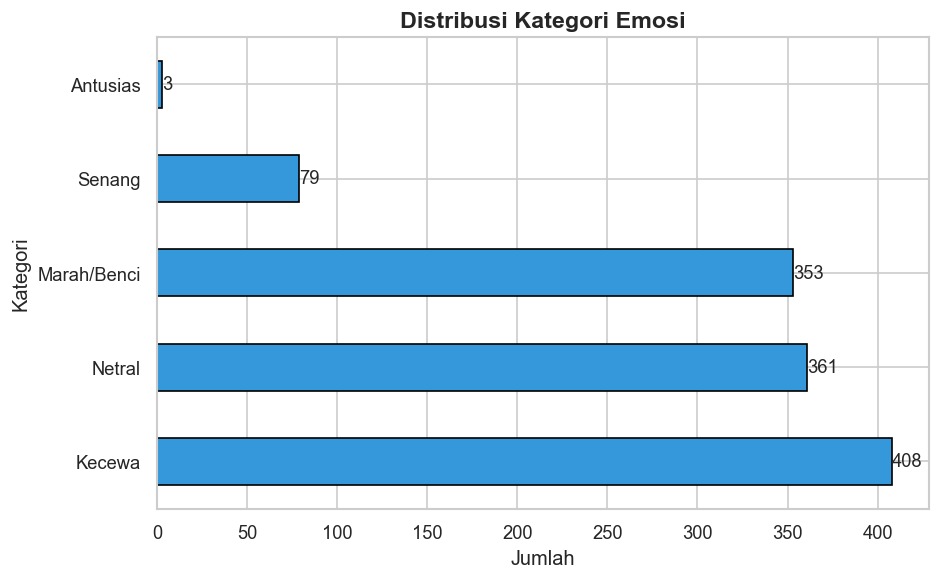} 
    \captionof{figure}{Statistical Spread of Text Length}
    \label{fig:text_length}
\end{center}
\vspace{0.2cm}

\subsubsection{Lexical and Semantic Analysis}
Extracting high-frequency unigrams and bigrams uncovers an unexpected phenomenon: the YouTube comments are profoundly saturated with socio-political rhetoric rather than typical service complaints.

Negative subsets are heavily plagued by conspiratorial jargon such as \textit{"antek asing"} (foreign minions), and \textit{"george soros"}. Conversely, affirmative sentiments frequently encompass patriotic or appreciative phrases like \textit{"terima kasih"} (thank you). This spillover of political discourse into commercial platforms highlights the complex interconnectedness of netizen behavior in Indonesia.

\vspace{0.2cm}
\begin{center}
    \includegraphics[width=\columnwidth]{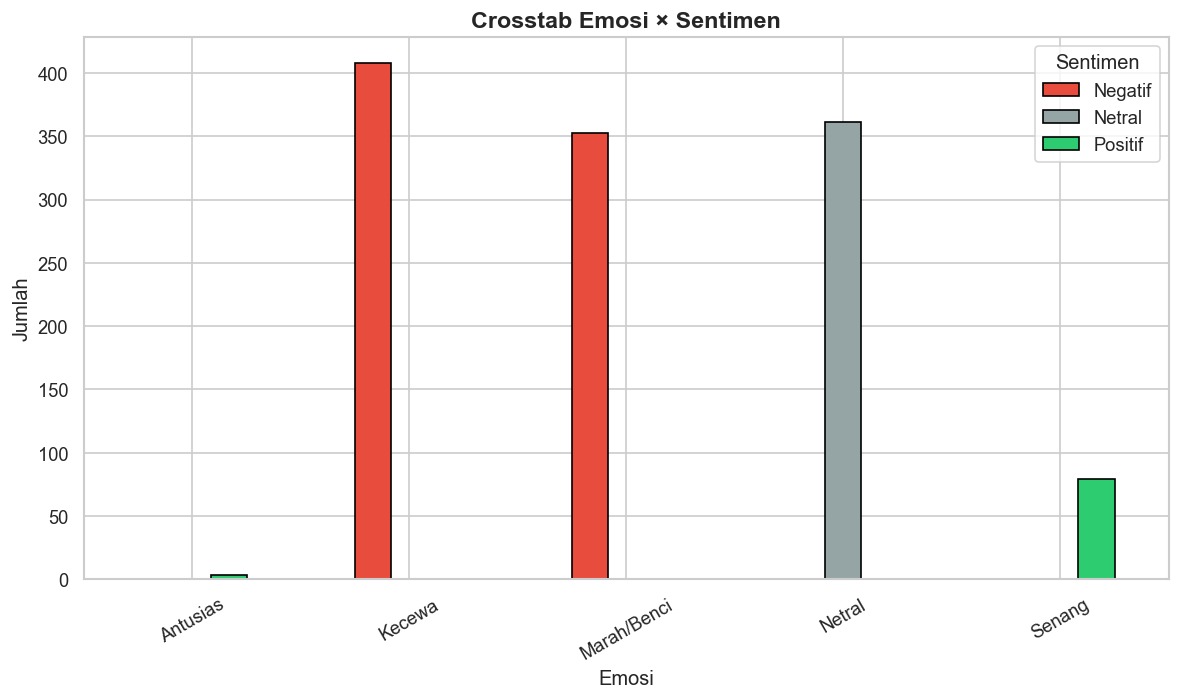}\\
    \vspace{0.1cm}
    \includegraphics[width=\columnwidth]{gambar5.jpeg}
    \captionof{figure}{High-Frequency Unigrams (Top) and Bigrams (Bottom)}
    \label{fig:ngrams_combined}
\end{center}
\vspace{0.2cm}

\subsubsection{Word Cloud Visualizations}
Semantic mapping via word clouds visually confirms the statistical observations. The aggregated visualization emphasizes words like "asing", "antek", and "Soros". When dissected by emotional categories, the divergence in vocabulary usage between aggressive negativity and supportive positivity becomes distinctly apparent.

\vspace{0.2cm}
\begin{center}
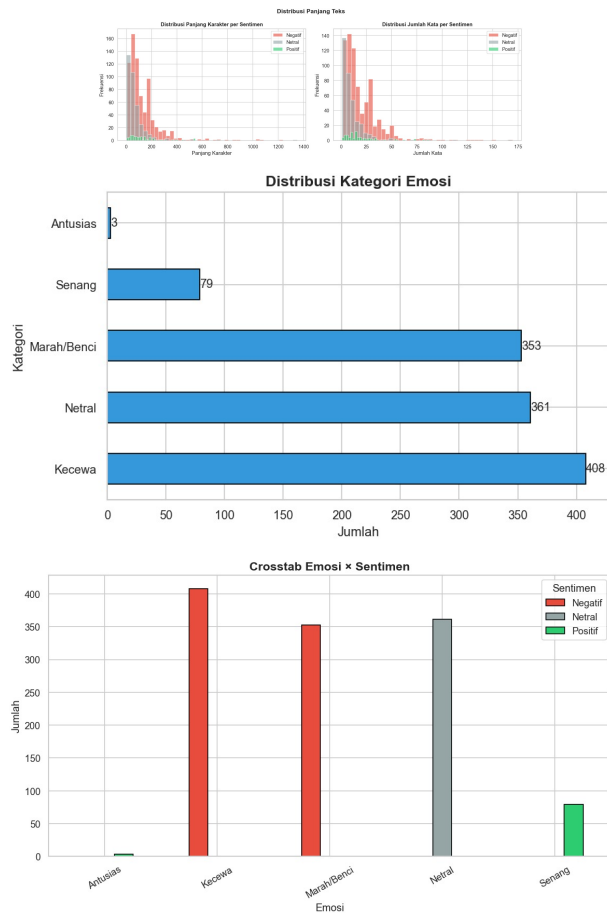

    \includegraphics[width=0.7\columnwidth]{gambar4.jpeg}\\
    \vspace{0.1cm}
    \includegraphics[width=\columnwidth]{gambar3.jpeg}\\
    \vspace{0.1cm}
    \includegraphics[width=\columnwidth]{gambar2.jpeg}
    \captionof{figure}{Corpus Word Clouds: Aggregated (Top), by Sentiment (Middle), by Emotion (Bottom)}
    \label{fig:wc_combined}
\end{center}
\vspace{0.2cm}

\subsection{Preprocessing Efficacy}
Raw digital interactions are intrinsically flawed by non-alphabetic characters. The preprocessing pipeline effectively eradicated emojis, punctuation, and capitalizations, transforming the dataset into a standardized format suitable for mathematical modeling (Figure \ref{fig:preprocessing_table}). 

\vspace{0.2cm}
\begin{center}
    \includegraphics[width=\columnwidth]{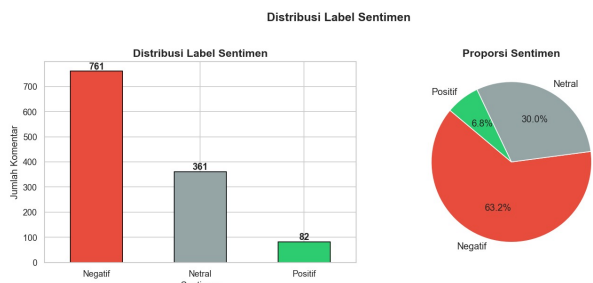}
    \captionof{figure}{Text Transformation Comparison}
    \label{fig:preprocessing_table}
\end{center}
\vspace{0.2cm}

A secondary word cloud constructed purely from the cleansed text ensures that critical contextual markers were preserved while grammatical noise was successfully discarded.

\vspace{0.2cm}
\begin{center}
    \includegraphics[width=\columnwidth]{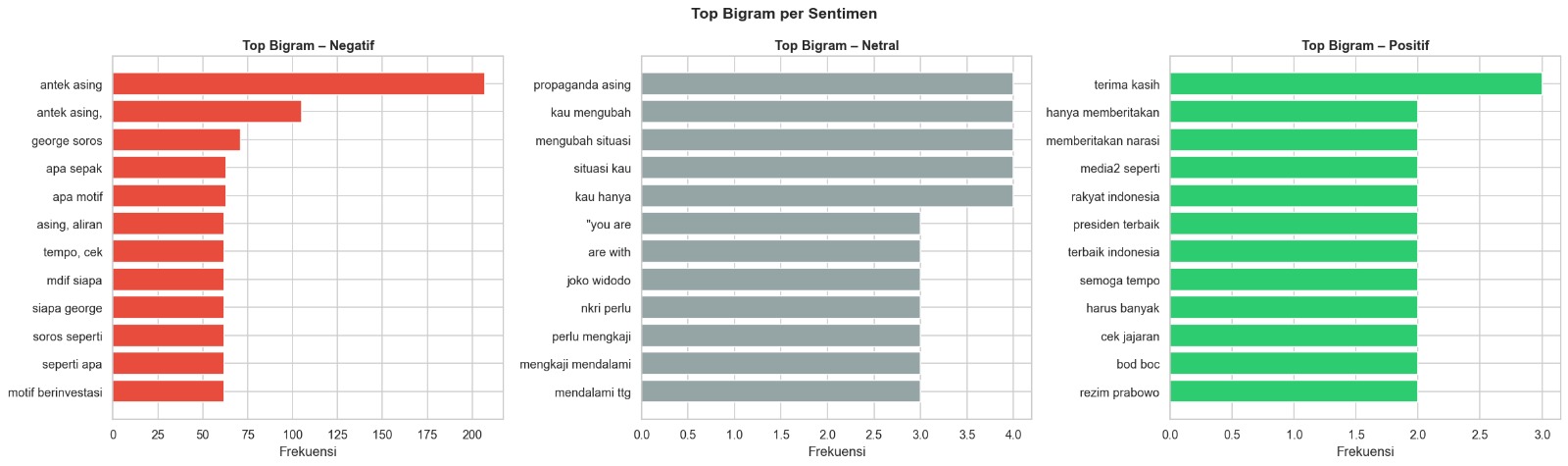}
    \captionof{figure}{Verification Word Cloud Post-Cleansing}
    \label{fig:wc_clean}
\end{center}
\vspace{0.2cm}

\subsection{Predictive Performance Benchmarking}

\subsubsection{Machine Learning Ensembles}
To identify the premier classifier, various models were rigorously tested. Support Vector Machines (SVM) initially secured the highest traditional baseline (Accuracy 76\%, F1-Score 72\%). An Automated Machine Learning (AutoML) architecture powered by PyCaret further refined these parameters, maintaining the optimal peak of 76\% accuracy.

\vspace{0.2cm}
\begin{center}
    \includegraphics[width=0.9\columnwidth]{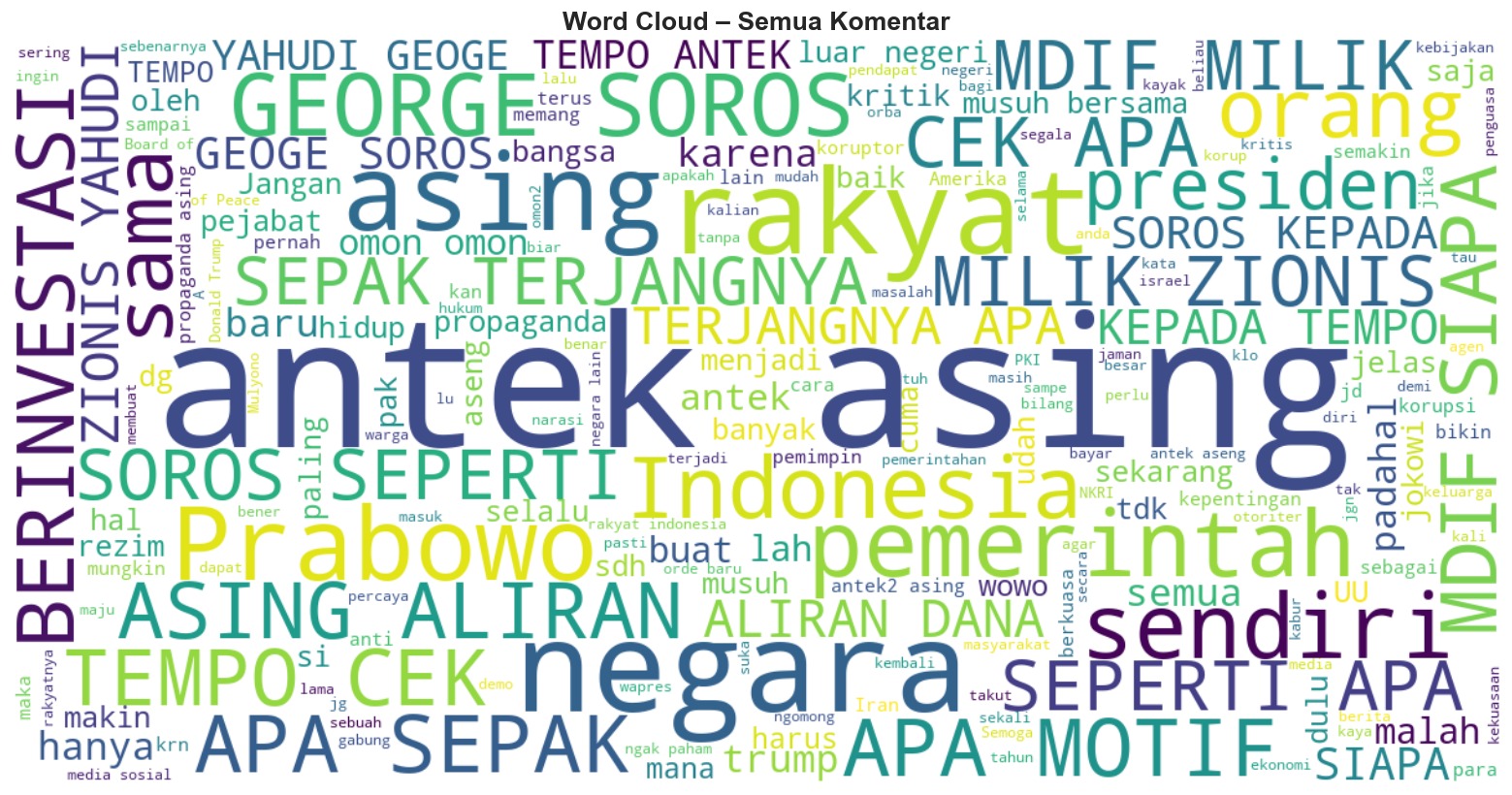}\\
    \vspace{0.1cm}
    \includegraphics[width=\columnwidth]{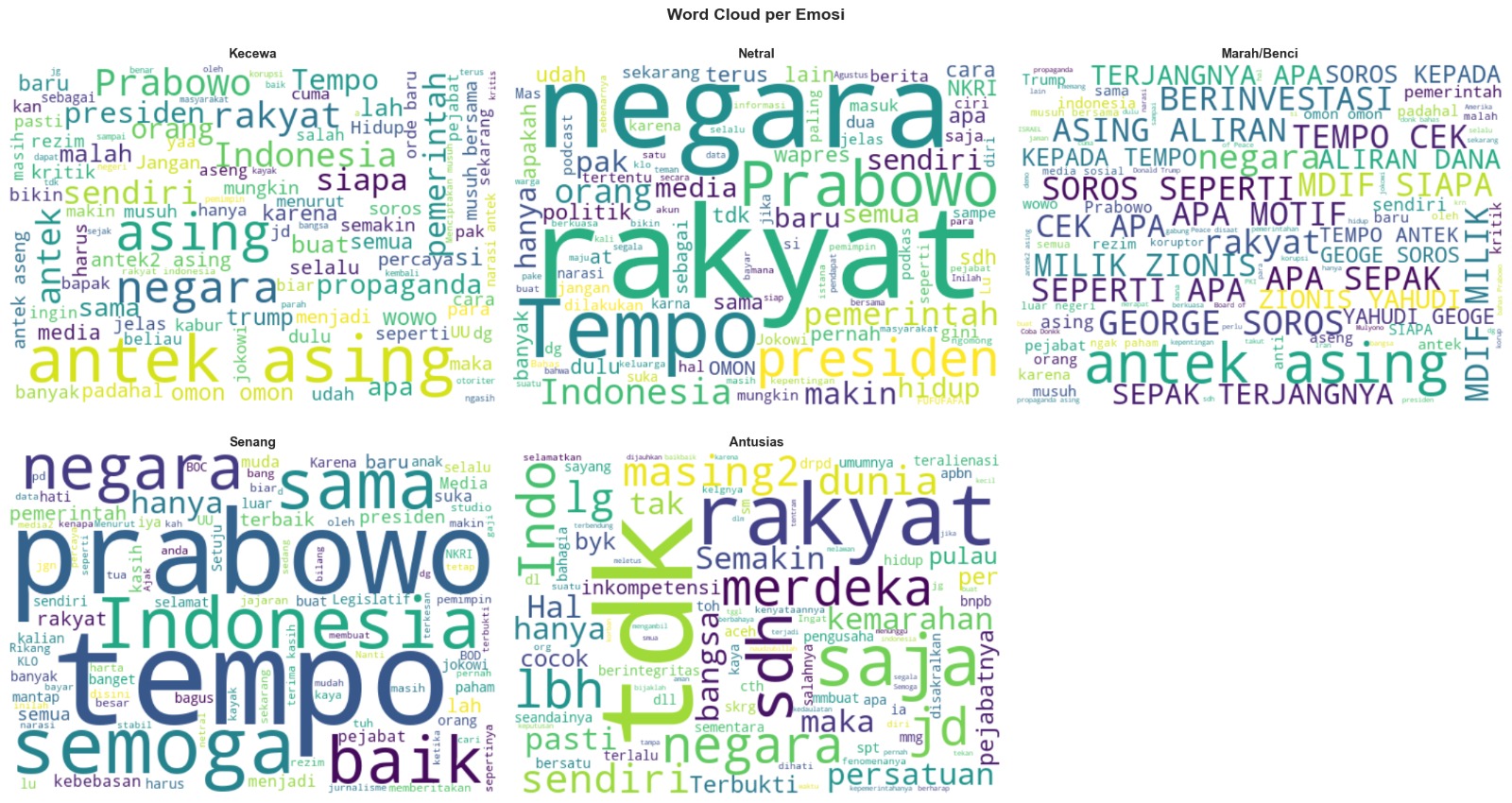}
    \captionof{figure}{Algorithm Benchmarks (Top) and Optimized ML Metrics (Bottom)}
    \label{fig:ml_combined}
\end{center}
\vspace{0.2cm}

\subsubsection{Comparative Evaluation: ML vs. LSTM}
Given the chronological sequence inherent in textual data, an LSTM neural network was deployed. The LSTM successfully classified 131 True Negatives, 29 True Neutrals, and 5 True Positives. Intriguingly, the traditional PyCaret-optimized Machine Learning approach (76\%) marginally eclipsed the LSTM architecture (74\%). 

This occurrence is highly attributable to the dataset's constrained volume and stark class imbalance. Traditional boosted algorithms process sparse TF-IDF matrices far more efficiently on moderate datasets, whereas deep learning frameworks typically require massive data influxes to surpass classical ensemble methods.

\vspace{0.2cm}
\begin{center}
    \includegraphics[width=0.8\columnwidth]{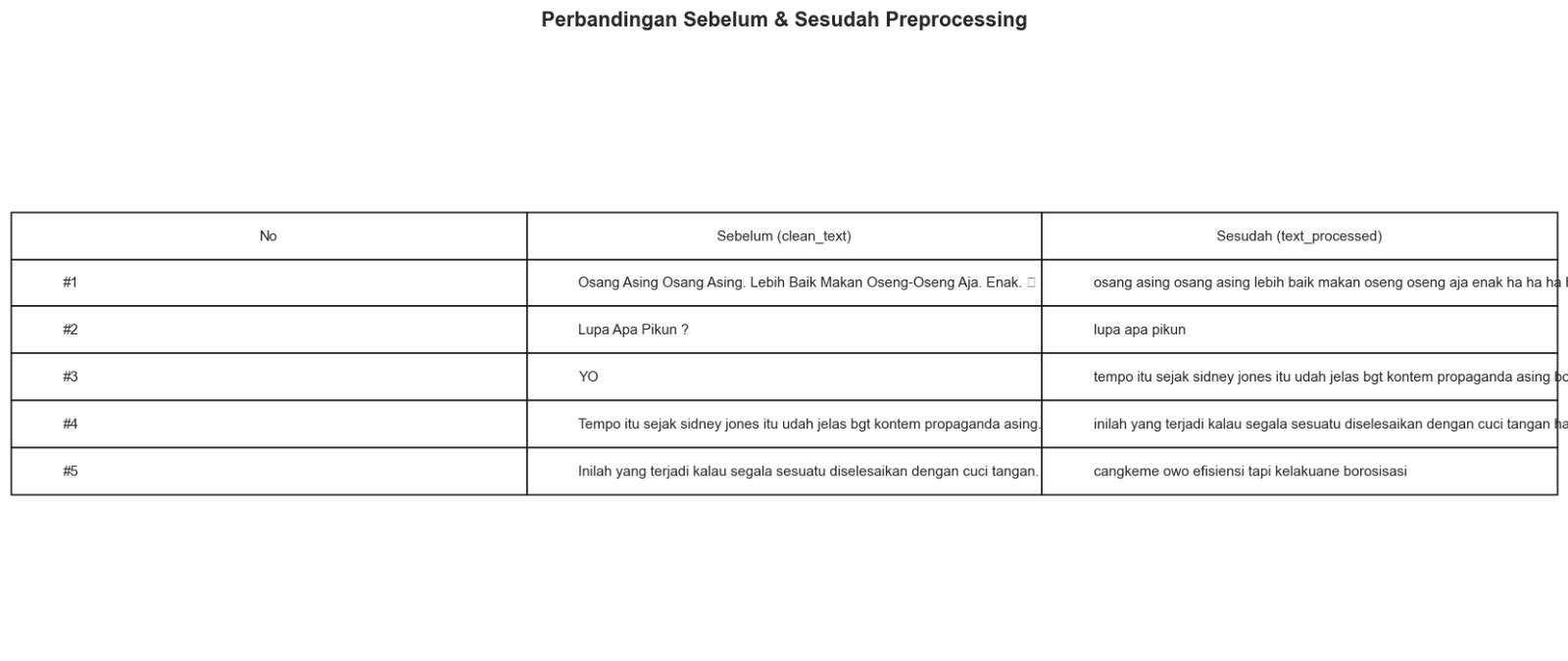}\\
    \vspace{0.1cm}
    \includegraphics[width=0.9\columnwidth]{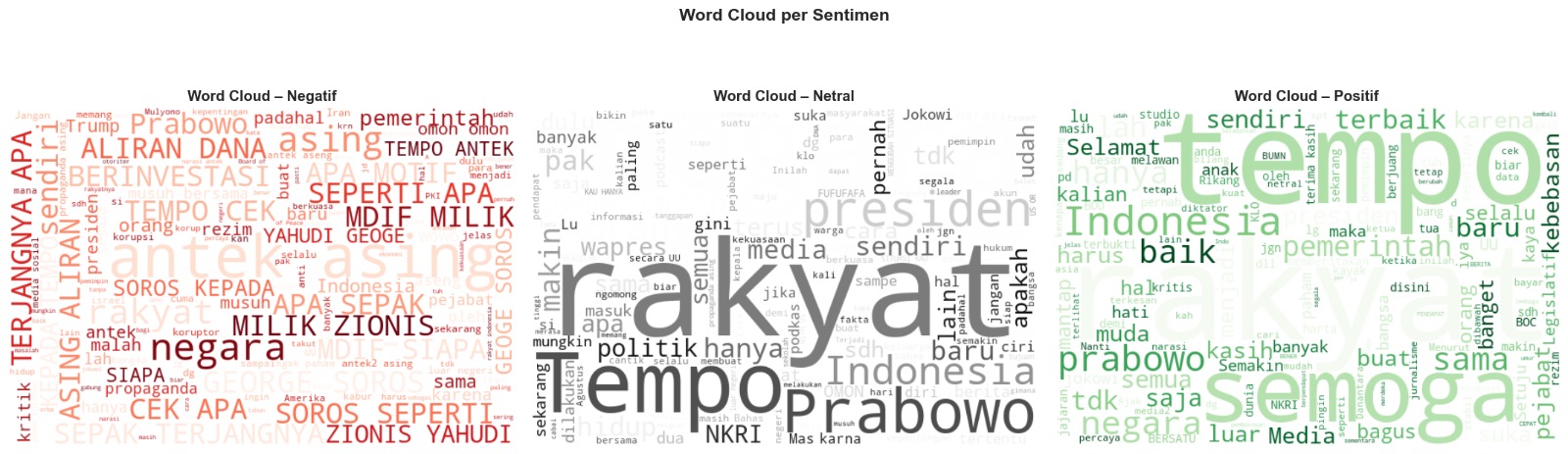}
    \captionof{figure}{LSTM Confusion Matrix (Top) and Model Confrontation (Bottom)}
    \label{fig:lstm_vs_ml}
\end{center}
\vspace{0.2cm}

% ── 5. CONCLUSION AND RECOMMENDATIONS ──────────────────────────────────────────
\section{Conclusion and Recommendations}

\subsection{Conclusion}
This research unequivocally demonstrates that optimized traditional Machine Learning architectures, notably those utilizing XGBoost and automated frameworks, offer highly pragmatic and accurate solutions for classifying noisy social media sentiments. The application of TF-IDF reliably captured the underlying textual semantics. Crucially, the exploratory evaluation highlighted a significant infiltration of socio-political discourse within commercial e-commerce reviews. Furthermore, the empirical evidence proved that on moderately sized, imbalanced datasets, ensemble ML models (76\%) possess the capability to slightly outperform resource-intensive Deep Learning LSTM networks (74\%).

\subsection{Recommendations}
To counteract the inherent skewness of public opinion data, future researchers are strongly advised to incorporate synthetic oversampling methodologies such as SMOTE. Expanding the corpus volume exponentially will also be vital to fully unlocking the long-term contextual learning potential of Deep Learning mechanisms in detecting nuanced sarcasm.

% ── REFERENCES ─────────────────────────────────────────────────────────────────
\renewcommand{\refname}{References}

\end{multicols}
\end{document}